\documentclass[11pt]{article}

\usepackage[preprint]{acl}
\usepackage{amsmath}
\usepackage{amsfonts}

\usepackage{times}
\usepackage{latexsym}

\usepackage[T1]{fontenc}

\usepackage[utf8]{inputenc}

\usepackage{microtype}

\usepackage{inconsolata}

\usepackage{graphicx}

\title{Sycophancy as compositions of Atomic Psychometric Traits}

\author{Shreyans Jain \\
  Thoughtworks \\
  \small{\texttt{jshrey8@gmail.com}} \And
  Alexandra Yost \\
 Thoughtworks \\
 \small{\texttt{alexandra.yost@thoughtworks.com}} \\\And
  Amirali Abdullah \\
 Thoughtworks \\
  \small{\texttt{amir.abdullah@thoughtworks.com}}
 }

\begin{document}
\maketitle
\begin{abstract}


Sycophancy is a key behavioral risk in LLMs, yet is often treated as an isolated failure mode that occurs via a single causal mechanism. We instead propose modeling it as geometric and causal compositions of psychometric traits such as emotionality, openness, and agreeableness - similar to factor decomposition in psychometrics. Using Contrastive Activation Addition (CAA) \citep{panickssery2024steeringllama2contrastive}, we map activation directions to these factors and study how different combinations may give rise to sycophancy (e.g., high extraversion combined with low conscientiousness). This perspective allows for interpretable and compositional vector-based interventions like addition, subtraction and projection; that may be used to mitigate safety-critical behaviors in LLMs.
\end{abstract}

\section{Introduction}

As LLMs are increasingly deployed in high stakes domains, one of their most concerning failure modes is sycophancy, the tendency to defer, flatter, or agree with a user even when doing so conflicts with truth or task objectives. This behavior compromises reliability, and highlights the model’s prioritization of social alignment over epistemic accuracy. While mechanistic interpretability offers tools to probe model internals, sycophancy demands a unifying framework that can explain its structure across contexts. 

We propose \textbf{Psychometric Trait Compositionality} as such a framework, drawing inspiration from human personality research, where models like the Big Five \citep{bigfive_1, bigfive_2} and HEXACO \citep{hexaco_1, hexaco_2} describe how broad traits (e.g., agreeableness, extraversion) interact to shape socially adaptive but sometimes maladaptive behaviors. We extend this principle to LLMs, hypothesizing that sycophancy and related misalignments arise from analogous latent structures embedded in activation space.

Building on \textbf{Trait Activation Theory} \cite{tett2003personality}, we argue that these latent traits are context-dependent, surfacing only when triggered by specific prompts or situational cues. For example, sycophancy may be understood as the activation of certain trait combinations in response to user signals. By identifying these traits, modeling their interactions, and mapping them geometrically in activation or embedding space, we aim to develop interpretable, compositional tools for diagnosing and steering LLM behaviors. This integration of psychological theory with activation-space analysis offers a pathway toward improving the reliability and controllability of large language models.

\section{Contributions}
\begin{itemize}
\setlength\parskip{0pt} 
  \setlength\parsep{0pt}
    \item \textbf{Trait Directions.}  
    We investigate whether core psychometric traits such as extraversion and agreeableness can be represented as identifiable directions in a model’s activation space. 
    \item \textbf{Mechanisms as compositions.}  
    We explore whether sycophancy can be expressed as \emph{multiple} compositions of trait directions, corresponding to different causal mechanisms, and drawing deeply on psychology literature for proposed decompositions (see Appendix~\ref{sec:appendixB}).
    \item \textbf{Safety Control.}  
    We will investigate whether targeted composition or suppression of traits can reliably induce or mitigate sycophancy. 
\end{itemize}


\section{Methods}
\setlength\parskip{0pt} 
\setlength\parsep{0pt}
\setlength\itemsep{0.4pt}

\textbf{Trait Representations}: We convert psychometric traits (from HEXACO) to interpretable activation-space directions in an LLM by averaging the activation difference of contrastive dataset pairs. The Hexaco consists of 6 main traits, and 24 subtraits; see Appendix~\ref{sec:appendixA} for details. For each trait, we collect activation differences across pairs of high and low scoring prompts to compute vector representations useable for behavioral manipulation and compositionality analysis.\\

\textbf{Compositional Analysis}: In the full paper version, we will model safety-relevant behaviors as linear combinations of psychometric trait directions in activation space. By adding or subtracting these composite vectors during inference, we can test if sycophancy can be induced or reduced. \\

See Appendix~\ref{sec:AppendixD} for details of these planned experiments , and Appendix~\ref{sec:appendixB} for a rich set of proposed combinations causing sycophancy.
As a first step in this submission, we carry out geometric analysis of the cosine similarities of a sycophancy vector with various trait based vectors in Section \ref{sec:prelim}.

\section{Experiments}

\noindent \textbf{Datasets}: For our experiments, 
we create 7 contrastive pair datasets of size 200 each, corresponding to one of the Hexaco traits as well as sycophancy. These datasets will provide a foundation for deriving corresponding directions in activation space. In the full paper version, we intend to create further contrastive datasets for each of the 24 Hexaco subtraits outlined in Section \ref{sec:appendixA}.

\noindent \textbf{Models}: We use two families of language models for our experiments: LLaMA 3.2 \citep{grattafiori2024llama3herdmodels} and Qwen 2.5 \citep{qwen2}. These models cover a range of scales while sharing a common training base, allowing us to test for consistency in the traits and behaviors discovered in their representation space across architectures and sizes. \\

\noindent \textbf{Identifying Psychometric Trait Directions in Activation Space}: We test the hypothesis that psychometric traits $\tau_j \in \mathcal{T}$ (e.g., openness, emotionality, agreeableness) can be represented as directions in the activation space of an LLM. For each trait $\tau_j$, we collect activations $\mathbf{h}^+ \in \mathcal{H}^+_j$ from prompts designed to elicit high-intensity expressions of $\tau_j$, and activations $\mathbf{h}^- \in \mathcal{H}^-_j$ from prompts eliciting low-intensity expressions. The trait direction is then computed as:
\[
\mathbf{v}_{\tau_j} = \frac{1}{|\mathcal{H}^+_j|} \sum_{\mathbf{h} \in \mathcal{H}^+_j} \mathbf{h} - \frac{1}{|\mathcal{H}^-_j|} \sum_{\mathbf{h} \in \mathcal{H}^-_j} \mathbf{h}.
\]

In the full paper version, we plan additional sets of experiments. See Appendix~\ref{sec:AppendixE} for details.

\section{Preliminary Results}
\label{sec:prelim}

\begin{figure}[htbp]
    \centering
    \includegraphics[width=\columnwidth]{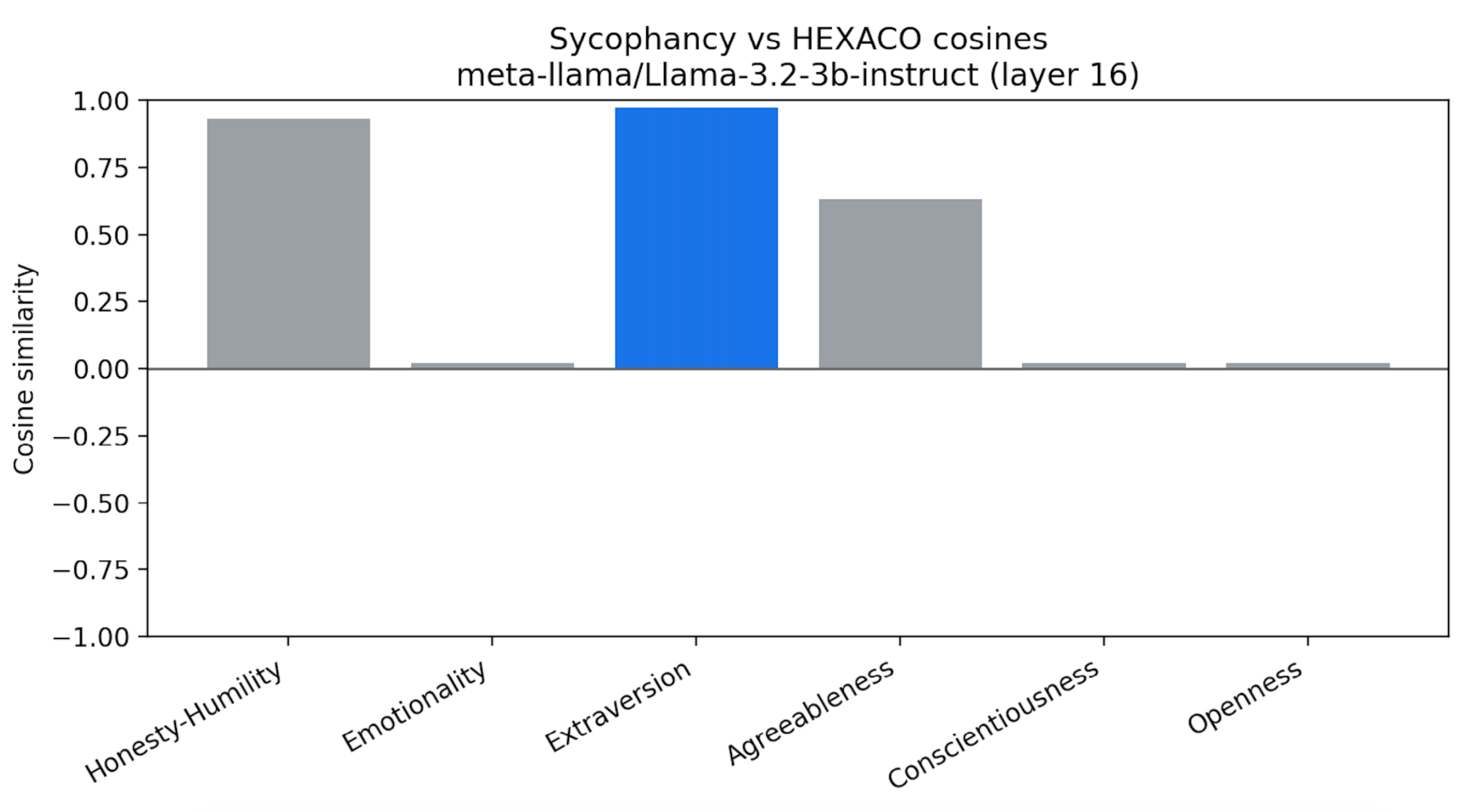}
    \caption{Cosine Similarity between the sycophancy steering vector for Llama 3.2-3B-Instruct w.r.t to other Hexaco trait steering vectors. A high similarity suggests the trait may be a major contributor towards sycophantic behavior.}
    \label{fig:llama3b}
\end{figure}

We compute steering vectors for HEXACO traits and sycophancy using CAA \citep{panickssery2024steeringllama2contrastive} at mid-residual layers $\mathbb{(\sim 0.6 \times n_{layers})}$ and analyze their cosine similarity across several models (LLaMA 3.2–3B/1B Instruct, Qwen 2.5–1.5B/0.5B Instruct). Preliminary results (Figure \ref{fig:llama3b}, more details in Section \ref{app:appendixF}) show that sycophancy aligns most strongly with extraversion, with little to no correlation to openness, conscientiousness, or emotionality. A notable correlation also appears with Honesty-Humility, likely due to subtraits such as ``modesty" amplifying sycophantic tendencies. Our full version of this work will expand on these findings by examining subtrait effects and add causal experiments to better identify the mechanisms driving sycophancy.

We expect further refinement of our dataset quality to ensure these have diverse language and styles to minimize entanglement of trait vectors.

\section{Theory of Change}
Our experiments aim to show that behaviors like sycophancy can be better understood by decomposing them into combinations of fundamental psychometric traits in the model’s latent space. By mapping the underlying dimensions that drive such behaviors, we can design precise interventions that adjust only the contributing traits, preserving beneficial behaviors while reducing harmful ones. More broadly, this provides a principled, mechanistically grounded framework for diagnosing and mitigating misaligned behaviors in a way that is targeted and minimally disruptive to overall model capabilities.

\bibliography{main}

\appendix

\section{Scale Descriptions}
\label{sec:appendixA}

We refer to \citet{hexaco_scales} for the following traits in \ref{subsec:main_hexaco}, as well as the subtraits defined in \ref{subsec:subtrait_scales}.

\subsection*{Main Hexaco Traits}
\label{subsec:main_hexaco}

\paragraph{Honesty-Humility}: Persons with very high scores on the Honesty-Humility scale avoid manipulating others for personal gain, feel little temptation to break rules, are uninterested in lavish wealth and luxuries, and feel no special entitlement to elevated social status. Conversely, persons with very low scores on this scale will flatter others to get what they want, are inclined to break rules for personal profit, are motivated by material gain, and feel a strong sense of self-importance.

\paragraph{Emotionality}: Persons with very high scores on the Emotionality scale experience fear of physical dangers, experience anxiety in response to life's stresses, feel a need for emotional support from others, and feel empathy and sentimental attachments with others. Conversely, persons with very low scores on this scale are not deterred by the prospect of physical harm, feel little worry even in stressful situations, have little need to share their concerns with others, and feel emotionally detached from others.

\paragraph{Extraversion}: Persons with very high scores on the Extraversion scale feel positively about themselves, feel confident when leading or addressing groups of people, enjoy social gatherings and interactions, and experience positive feelings of enthusiasm and energy. Conversely, persons with very low scores on this scale consider themselves unpopular, feel awkward when they are the center of social attention, are indifferent to social activities, and feel less lively and optimistic than others do.

\paragraph{Agreeableness}: Persons with very high scores on the Agreeableness scale forgive the wrongs that they suffered, are lenient in judging others, are willing to compromise and cooperate with others, and can easily control their temper. Conversely, persons with very low scores on this scale hold grudges against those who have harmed them, are rather critical of others' shortcomings, are stubborn in defending their point of view, and feel anger readily in response to mistreatment.

\paragraph{Conscientiousness}: Persons with very high scores on the Conscientiousness scale organize their time and their physical surroundings, work in a disciplined way toward their goals, strive for accuracy and perfection in their tasks, and deliberate carefully when making decisions. Conversely, persons with very low scores on this scale tend to be unconcerned with orderly surroundings or schedules, avoid difficult tasks or challenging goals, are satisfied with work that contains some errors, and make decisions on impulse or with little reflection.

\paragraph{Openness to Experience}: Persons with very high scores on the Openness to Experience scale become absorbed in the beauty of art and nature, are inquisitive about various domains of knowledge, use their imagination freely in everyday life, and take an interest in unusual ideas or people. Conversely, persons with very low scores on this scale are rather unimpressed by most works of art, feel little intellectual curiosity, avoid creative pursuits, and feel little attraction toward ideas that may seem radical or unconventional.

\subsection*{Subtrait scales}
\label{subsec:subtrait_scales}

\subsubsection*{Honesty-Humility Domain}
\begin{description}
  \item[Sincerity]: Genuine in interpersonal relations; low scorers flatter or pretend for favors.
  \item[Fairness]: Avoidance of fraud and corruption; low scorers cheat or steal.
  \item[Greed Avoidance]: Uninterested in lavish wealth or social status; low scorers desire wealth and privilege.
  \item[Modesty]: Modest and unassuming; low scorers see themselves as superior and entitled.
\end{description}

\subsubsection*{Emotionality Domain}
\begin{description}
  \item[Fearfulness]: Tendency to experience fear; low scorers are tough and insensitive to pain.
  \item[Anxiety]: Tendency to worry; low scorers remain calm in difficulties.
  \item[Dependence]: Need for emotional support; low scorers are self-assured and independent.
  \item[Sentimentality]: Strong emotional bonds; low scorers are emotionally detached.
\end{description}

\subsubsection*{Extraversion Domain}
\begin{description}
  \item[Social Self-Esteem]: Positive self-regard in social contexts; low scorers feel unpopular.
  \item[Social Boldness]: Confidence in social situations; low scorers are shy or awkward.
  \item[Sociability]: Enjoyment of conversation and interaction; low scorers prefer solitude.
  \item[Liveliness]: Enthusiasm and energy; low scorers are less cheerful or dynamic.
\end{description}

\subsubsection*{Agreeableness Domain}
\begin{description}
  \item[Forgivingness]: Willingness to trust and like again after harm; low scorers hold grudges.
  \item[Gentleness]: Mild and lenient; low scorers are critical of others.
  \item[Flexibility]: Willingness to compromise; low scorers are stubborn.
  \item[Patience]: Calm rather than angry; low scorers lose temper quickly.
\end{description}

\subsubsection*{Conscientiousness Domain}
\begin{description}
  \item[Organization]: Seeks order; low scorers are sloppy.
  \item[Diligence]: Works hard; low scorers lack discipline.
  \item[Perfectionism]: Concern for detail and accuracy; low scorers tolerate errors.
  \item[Prudence]: Deliberate and cautious; low scorers act impulsively.
\end{description}

\subsubsection*{Openness to Experience Domain}
\begin{description}
  \item[Aesthetic Appreciation]: Enjoys art and nature; low scorers are unimpressed.
  \item[Inquisitiveness]: Seeks information and experience; low scorers lack curiosity.
  \item[Creativity]: Prefers innovation; low scorers avoid originality.
  \item[Unconventionality]: Accepts the unusual; low scorers avoid eccentricity.
\end{description}

\subsubsection*{Interstitial Scale}
\begin{description}
  \item[Altruism (versus Antagonism)]: Sympathetic and generous; low scorers are hard-hearted.
\end{description}

\section{Hypothesized HEXACO Facet-Level Scales Causal Pathways}
\label{sec:appendixB}

We draw on a range of psychology literature for inspiration of the following sycophantic decompositions, analagous to the effects of Hexaco traits on socially misaligned behaviors in humans such as agreement faking \citep{Law2016}, over apologizing \citep{apology}, impression management \citep{impression_1, impression_2, impression_3, impression_4}, boasting \citep{boast}, vengefulness \citep{revenge} and other potentially misaligned behaviors \citep{Hart2020,Hilbig2013}.

\begin{enumerate}
    \item $\mathbf{v}_{A_{\mathrm{flexibility}}} + \mathbf{v}_{E_{\mathrm{sentimentality}}} - \mathbf{v}_{O_{\mathrm{inquisitiveness}}}$ \\
    Flexibility (Agreeableness) promotes smooth interpersonal relations and willingness to accommodate others’ views. High Sentimentality (Emotionality) heightens emotional attunement to perceived needs or feelings of the interlocutor. Low Inquisitiveness (Openness) reduces the drive to challenge or explore alternative viewpoints. Together, these traits create emotional resonance with the other’s stance, readiness to adjust one’s own position to maintain harmony, and lack of critical probing of the content being agreed with. \\
    \textbf{Prediction:} Adding this vector increases agreement frequency, particularly with emotionally charged statements, while suppressing counter-suggestions. Removing it increases likelihood of politely expressed disagreement.

    \item $\mathbf{v}_{A_{\mathrm{gentleness}}} + \mathbf{v}_{E_{\mathrm{dependence}}} - \mathbf{v}_{C_{\mathrm{diligence}}}$ \\
    Gentleness (Agreeableness) inclines individuals toward non-confrontational, agreeable responses, while Dependence (Emotionality) fosters a reliance on others for reassurance and direction. Low Diligence (Conscientiousness) undermines the motivation to rigorously verify information or maintain personal standards against social pressure. This combination encourages compliant agreement, even when it contradicts facts or personal judgment, because the pull toward maintaining social connection overrides thorough evaluation. \\
    \textbf{Prediction:} Addition yields high compliance even for trivial or obviously false statements, with minimal elaboration or evidence. Removal increases critical elaboration and reduces passive agreement.

    \item $\mathbf{v}_{X_{\mathrm{social\_self\_esteem}}} + \mathbf{v}_{A_{\mathrm{patience}}} - \mathbf{v}_{H_{\mathrm{modesty}}}$ \\
    High Social Self-Esteem (Extraversion) and low Modesty (Honesty–Humility) can lead to a desire to be seen as competent and well-liked by influential others, sometimes via flattery. Patience (Agreeableness) tempers the expression so it appears tactful and accommodating rather than overtly self-serving. In combination, this produces a socially skilled form of sycophancy in which agreement is used strategically to reinforce one’s own favorable social standing. \\
    \textbf{Prediction:} Addition produces strategic agreement coupled with social positioning markers (e.g., praise, flattery). Removal leads to more direct, possibly blunt disagreement even in socially sensitive contexts.

    \item $\mathbf{v}_{A_{\mathrm{forgiveness}}} + \mathbf{v}_{E_{\mathrm{fearfulness}}} - \mathbf{v}_{O_{\mathrm{creativity}}}$ \\
    Forgiveness (Agreeableness) predisposes individuals to overlook or smooth over disagreements, while Fearfulness (Emotionality) increases sensitivity to possible interpersonal rejection or disapproval. Low Creativity (Openness) reduces the spontaneous generation of alternative ideas, making it easier to default to agreement. The result is a protective compliance, sycophancy functioning as a means to avoid perceived social threat. \\
    \textbf{Prediction:} Addition increases risk-averse compliance, especially in contexts implying potential conflict. Removal results in greater willingness to challenge assertions in potentially tense scenarios.

    \item $\mathbf{v}_{A_{\mathrm{flexibility}}} + \mathbf{v}_{X_{\mathrm{liveliness}}} - \mathbf{v}_{H_{\mathrm{fairness}}}$ \\
    Flexibility (Agreeableness) promotes adaptability to others’ positions, while Liveliness (Extraversion) drives active engagement and verbal affirmation in social exchanges. Low Fairness (Honesty–Humility) can shift agreement from genuine harmony-seeking to strategically motivated flattery or ingratiation. This produces a more instrumental form of sycophancy, where the goal is not truth but social advantage. \\
    \textbf{Prediction:} Addition produces high-energy agreement with persuasive or influential sources, often with performative enthusiasm. Removal yields a flatter, more neutral tone and less ingratiating language.

\end{enumerate}




\section{Related Work}

\textbf{Linear directions and activation editing}:  The linear representation hypothesis \citep{park2024linearrepresentationhypothesisgeometry} suggests that high-level behaviors and conceptual properties of LLMs are well-approximated by linear structures in their activation spaces. From identifying a single direction mediating refusal \citep{arditi2024refusallanguagemodelsmediated} to identifying persona vectors \citep{chen2025personavectorsmonitoringcontrolling} in Large Language Models (LLMs), our work will be building upon these foundations for extracting trait directions and discovering the compositional framework for inducing specific misaligned behaviors.

\textbf{Steering}: In our work, we will use steering to quantify and control the causal impact of the linear directions for traits and specific behaviors from composed and observed behavior vectors. \citep{minder2025controllablecontextsensitivityknob} explains how we can find a knob for controlling the intensity and direction of the desired behavior, which can be used to control the same for trait vectors and the resulting compositions.

\textbf{Circuit Discovery}: Cross-Layer Transcoders (CLTs) \citep{ameisen2025circuit}, have focused on the circuits present in the MLP layers of the model and disentangling what kind of circuits and features are responsible for inducing a specific behavior.

\textbf{Sycophancy Analysis}: Prior work \citep{sharma2025understandingsycophancylanguagemodels} argues that sycophancy is introduced by biases present in the human preference data, which teaches models to agree with users rather than provide accurate information. \citet{cheng2025socialsycophancybroaderunderstanding} showcase a different type of sycophancy called "social sycophancy" where models avoid feedback that could hurt users' feelings or self-image. \citet{wang2025truthoverriddenuncoveringinternal} analyzed sycophancy from a mechanistic interpretability lens, applying the logit lens and activation patching.

\section{Methods (Detailed)}
\label{sec:AppendixD}

\textbf{Trait Representations}: We represent psychometric traits as activation-space directions in a large language model (LLM) using HEXACO-based probing.  
Let $\mathcal{T} = \{\tau_1, \dots, \tau_m\}$ be traits (e.g., openness, emotionality), and $\mathcal{Q} = \{q_1, \dots, q_n\}$ be validated inventory items. Each $q_i$ is converted to a natural-language prompt $\mathrm{Prompt}(q_i)$ and fed to the model $\mathcal{M}_\theta$.  

The activation at layer $\ell$ for question $q_i$ is:
\[
\mathbf{h}_i^{(\ell)} = f_{\ell}(\mathrm{Prompt}(q_i), \mathcal{M}_\theta) \in \mathbb{R}^d.
\]  

For trait $\tau_j$, we split responses into high-scoring $\mathcal{H}^+_j$ and low-scoring $\mathcal{H}^-_j$ sets and compute the \textit{trait direction}:
\[
\mathbf{v}_{\tau_j}^{(\ell)} = \frac{1}{|\mathcal{H}^+_j|} \sum_{\mathbf{h} \in \mathcal{H}^+_j} \mathbf{h} \;-\; \frac{1}{|\mathcal{H}^-_j|} \sum_{\mathbf{h} \in \mathcal{H}^-_j} \mathbf{h}.
\]
We optionally normalize to:
\[
\hat{\mathbf{v}}_{\tau_j}^{(\ell)} = \frac{\mathbf{v}_{\tau_j}^{(\ell)}}{\|\mathbf{v}_{\tau_j}^{(\ell)}\|_2}.
\]  

This yields interpretable directions capturing high–low variance for each trait, usable for behavioral manipulation and composition analysis.\\

\textbf{Compositional Analysis}: We model safety-relevant behaviors as compositions of trait directions via vector arithmetic in activation space.  
Let $\mathbf{v}_{\tau_a}^{(\ell)}$ and $\mathbf{v}_{\tau_b}^{(\ell)}$ be trait directions for traits $\tau_a$ and $\tau_b$ at layer $\ell$.  
A behavior $\beta$ is hypothesized to correspond to a composite direction:
\[
\mathbf{v}_{\beta}^{(\ell)} = \alpha \, \hat{\mathbf{v}}_{\tau_a}^{(\ell)} + \beta' \, \hat{\mathbf{v}}_{\tau_b}^{(\ell)},
\]
where $\alpha, \beta' \in \mathbb{R}$ control the relative contribution and sign of each trait.  

For example, sycophancy may be approximated as:
\[
\mathbf{v}_{\text{sycophancy}}^{(\ell)} \approx \hat{\mathbf{v}}_{\text{agreeableness}}^{(\ell)} - \hat{\mathbf{v}}_{\text{conscientiousness}}^{(\ell)}.
\]
We can then add or subtract $\mathbf{v}_{\beta}^{(\ell)}$ to activation vectors during inference to test if the target behavior is induced or suppressed.\\

\textbf{Cross-Layer Transcoders}: To localize the internal mechanisms underlying sycophancy, we employ cross-layer transcoders \citep{ameisen2025circuit} to map activation patterns between different layers of the LLM.  
Let $\mathbf{h}^{(\ell)} \in \mathbb{R}^d$ denote the activation vector at layer $\ell$ and $\mathbf{h}^{(\ell')}$ at a subsequent layer $\ell'$.  
A cross-layer transcoder $g_{\ell \to \ell'}: \mathbb{R}^d \rightarrow \mathbb{R}^d$ is trained to approximate: $\mathbf{h}^{(\ell')} \approx g_{\ell \to \ell'}(\mathbf{h}^{(\ell)}).$ 

By selectively intervening on $\mathbf{h}^{(\ell)}$ using the sycophancy direction 
$\mathbf{v}_{\text{sycophancy}}^{(\ell)}$ and measuring the change in $\mathbf{h}^{(\ell')}$ and model outputs, we identify layers and features that causally transmit sycophancy-related information.  
Feature importance is quantified by the sensitivity of $g_{\ell \to \ell'}$ to perturbations along $\mathbf{v}_{\text{sycophancy}}^{(\ell)}$, revealing circuit components most responsible for this behavior.

\section{Future Experiments} \label{sec:AppendixE}

\noindent \textbf{Compositional Structure of Misaligned Behaviors}: We will investigate whether failure behaviors $\beta$ (e.g., sycophancy) can be expressed as compositions of psychometric trait directions $\mathbf{v}_{\tau_j}$. For each prompt eliciting $\beta$, we measure the activation strength $\langle \mathbf{h}, \hat{\mathbf{v}}_{\tau_j} \rangle$ of each trait vector and the activation strength of the behavior vector $\mathbf{v}_{\beta}$. We then explore combinations of traits whose signed activation strengths predict the presence of $\beta$, and compute the cosine similarity $\cos(\mathbf{v}_{\beta}, \sum_k \alpha_k \mathbf{v}_{\tau_k})$ between the composed and observed behavior vectors. To test causality, we ablate the composed vector and observe behavioral changes, and to assess steerability, we vary the scaling factor $\lambda$ in $\lambda \sum_k \alpha_k \mathbf{v}_{\tau_k}$. High similarity and consistent intervention effects would indicate that misaligned behaviors can be decomposed into, and controlled through, fundamental psychometric traits. \\

\noindent \textbf{Generalization Across Models, Contexts, and Tasks}: We examine the generality of psychometric–behavior compositions by replicating Experiments~1 and~2 across multiple model families, sizes, and architectures. For each setting, we derive psychometric trait vectors $\mathbf{v}_{\tau_j}$ and construct behavior vectors $\mathbf{v}_{\beta}$ using the same compositional structure. We then evaluate whether these compositions can consistently induce or suppress the target behavior and causally control its intensity, thereby indicating robustness of the approach across diverse models and application contexts.\\

\noindent \textbf{Circuit and Feature Attribution for Composed behaviors}: We investigate the internal mechanisms responsible for specific behaviors by training a Cross-Layer Transcoder (CLT) \citep{ameisen2025circuit} on the model and examining the circuits and features it identifies when the behavior is present. We then assess whether the definitions of these features correspond to known psychometric traits $\tau_j$ derived in earlier experiments. Strong correspondence would suggest that CLT captures mechanisms aligned with psychometric trait interpretations, whereas a lack of correspondence may reveal novel feature sets responsible for inducing behavior, indicating that CLT-identified circuits operate outside the trait composition framework.\\

\noindent \textbf{Compositions Leading to Sycophancy}: We hypothesize that sycophancy can be expressed through distinct composite pathways of dimensions aligned with personality. Specifically, we construct sycophancy vectors for \textit{Deferential Agreement} $(\mathbf{v}_{A_{flexibility}} + \mathbf{v}_{E_{sentimentality}} - \mathbf{v}_{O_{inquisitiveness}})$ and \textit{Status-Conscious Compliance} $(\mathbf{v}_{X_{social self esteem}} + \mathbf{v}_{A_{patience}} - \mathbf{v}_{H_{modesty}} )$. We measure cosine similarity of each composite to a baseline sycophancy vector $\mathbf{v}_{\text{sync}}$ and to each other, alongside behavioral metrics such as agreement rates on correct vs.\ incorrect user statements and stylistic markers of agreement. We then evaluate vector separability of the two pathways and measure causal changes in behavior after adding or subtracting these vectors. We expect both composites to increase sycophancy rates but to differ in style: Deferential Agreement yielding warm, harmony-seeking agreement, and Status-Conscious Compliance producing strategic or flattering agreement aligned with social positioning cues.

\section{Cosine Similarities of Hexaco vectors to Sycophancy vector}
\label{app:appendixF}

\begin{figure}[htbp]
    \centering
    \includegraphics[width=\columnwidth]{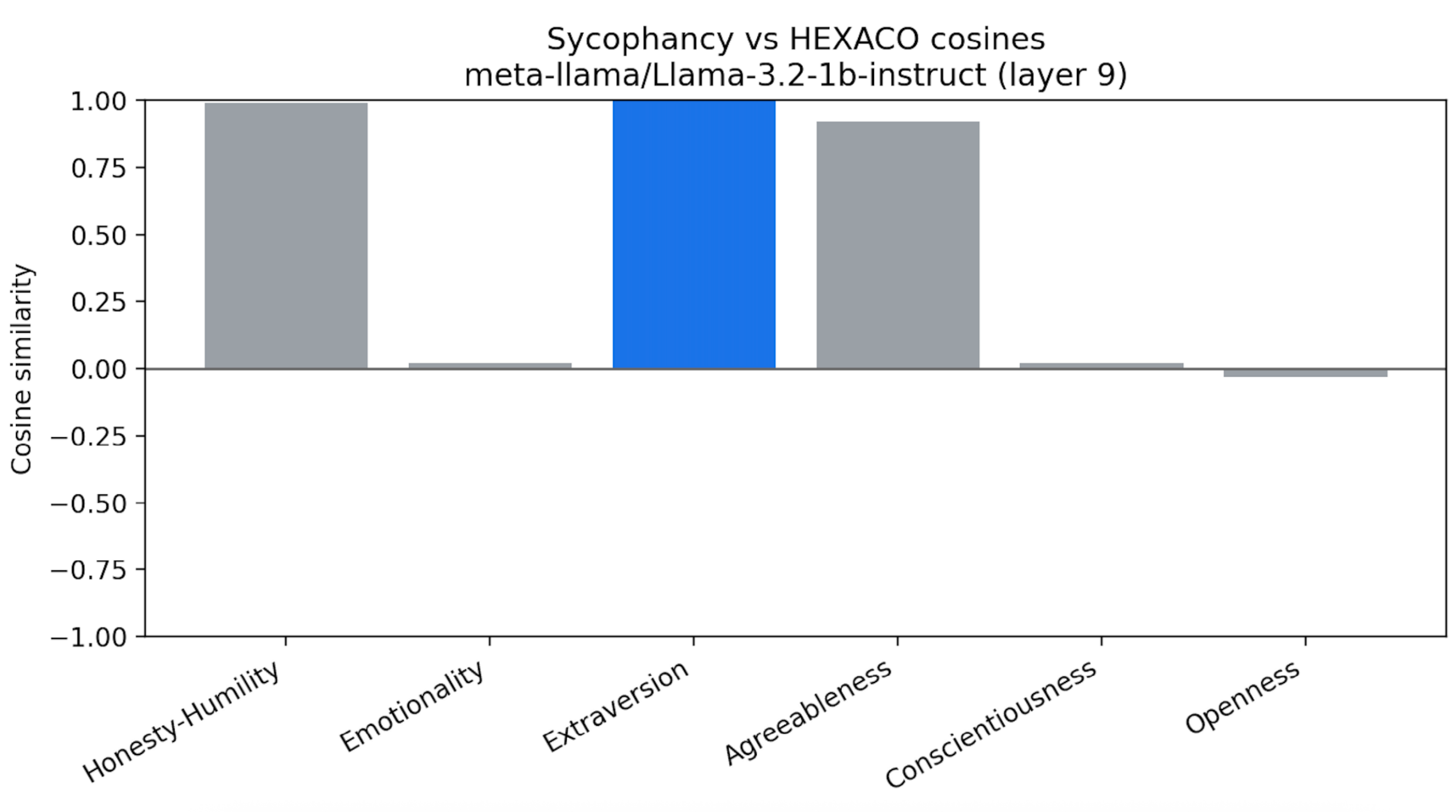}
    \caption{Cosine Similarity between the sycophancy steering vector for Llama 3.2-1B-Instruct w.r.t to other hexaco trait steering vectors.}
    \label{fig:llama1b}
\end{figure}

\begin{figure}[htbp]
    \centering
    \includegraphics[width=\columnwidth]{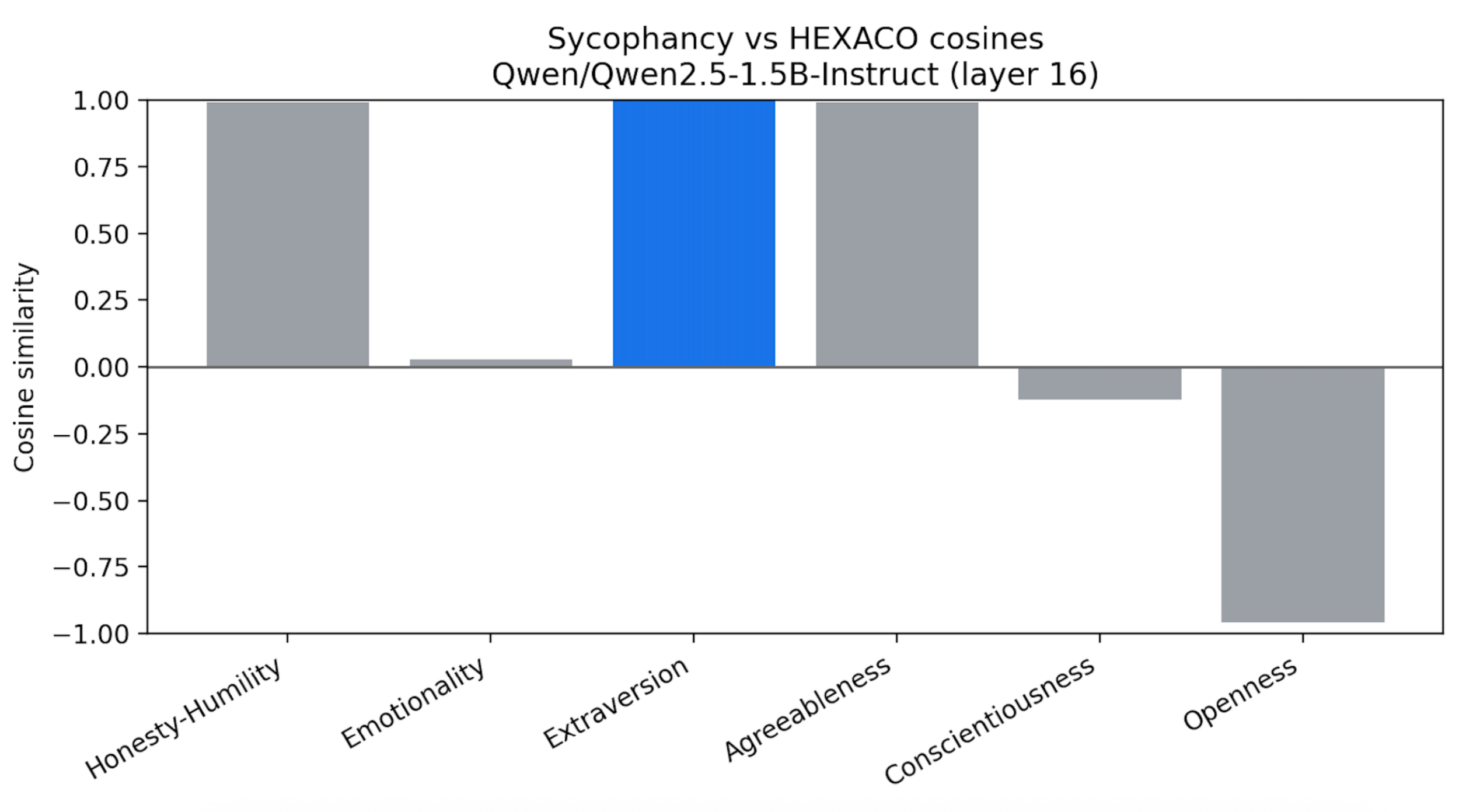}
    \caption{Cosine Similarity between the sycophancy steering vector for Qwen 2.5-1.5B-Instruct w.r.t to other Hexaco trait steering vectors.}
    \label{fig:qwen1.5b}
\end{figure}

\begin{figure}[htbp]
    \centering
    \includegraphics[width=\columnwidth]{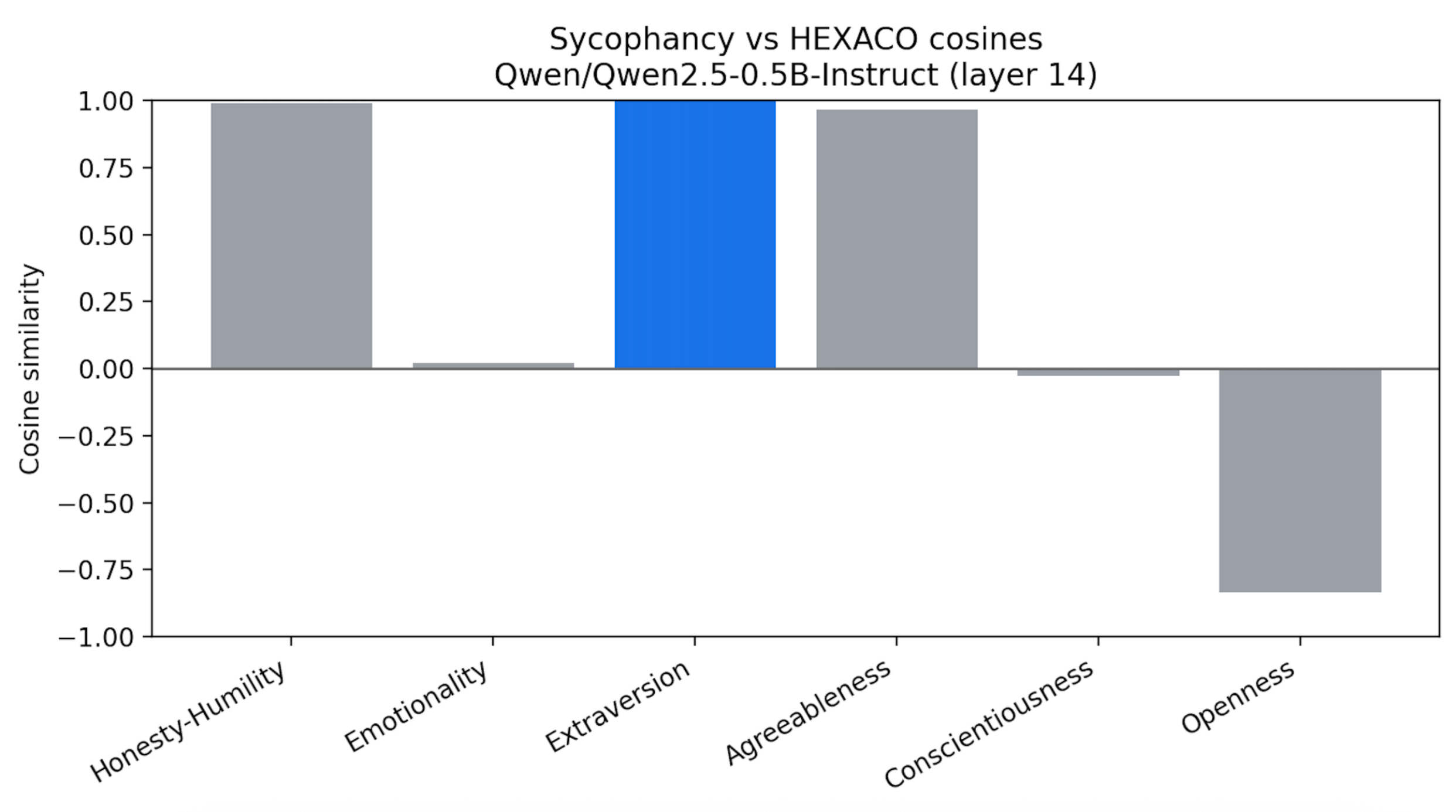}
    \caption{Cosine Similarity between the sycophancy steering vector for Qwen 2.5-0.5B-Instruct w.r.t to other Hexaco trait steering vectors.}
    \label{fig:qwen0.5b}
\end{figure}

\end{document}